\documentclass[envcountsame,oribibl]{llncs}
\usepackage{amssymb}
\usepackage{amsfonts}
\usepackage{amsmath}
\usepackage{dsfont} 
\usepackage{lipsum}
\usepackage{graphicx}

\usepackage{multirow}
\usepackage{stfloats}
\usepackage{floatrow}
\usepackage{algorithm}
\usepackage{algorithmic}
\usepackage{adjustbox}
\usepackage{bm}
\usepackage{booktabs}
\usepackage{threeparttable}
\makeatletter
\newcommand{\figcaption}{\def\@captype{figure}\caption}
\newcommand{\tabcaption}{\def\@captype{table}\caption}
\makeatother
\usepackage[misc]{ifsym}
\usepackage[tight,footnotesize]{subfigure}

\usepackage{colortbl}
\newcolumntype{+}{>{\global\let\currentrowstyle\relax}}
\newcolumntype{^}{>{\currentrowstyle}}

\newcommand{\PreserveBackslash}[1]{\let\temp=\\#1\let\\=\temp}
\newcolumntype{C}[1]{>{\PreserveBackslash\centering}p{#1}}
\newcolumntype{R}[1]{>{\PreserveBackslash\raggedleft}p{#1}}
\newcolumntype{L}[1]{>{\PreserveBackslash\raggedright}p{#1}}
\begin{document}
\title{Discriminative Consistent Domain Generation for Semi-supervised Learning}

\author{Jun Chen*\inst{1}, Heye Zhang*\inst{2}, Yanping Zhang\textsuperscript{1(\Letter)}, Shu Zhao\inst{1}, Raad Mohiaddin\inst{3,4}, Tom Wong\inst{3,4}, David Firmin\inst{3,4}, Guang Yang\textsuperscript{3,4(\Letter)}, Jennifer Keegan\inst{3,4}}

\institute{ $^1$School of Computer Science and Technology, Anhui University, Hefei, China \\
\email{zhangyp2@gmail.com}\\
$^2$School of Biomedical Engineering, Sun Yat-Sen University, Shenzhen, China\\
$^3$Cardiovascular Research Centre, Royal Brompton Hospital,
London SW3 6NP, UK \\
$^4$National Heart \& Lung Institute, Imperial College London,
London SW7 2AZ, UK\\
\email{g.yang@imperial.ac.uk}
}

\maketitle
\newcommand\blfootnote[1]{%
\begingroup
\renewcommand\thefootnote{}\footnote{#1}%
\addtocounter{footnote}{-1}%
\endgroup
}

\blfootnote{* These authors contributed equally to this work.}

\begin{abstract}
Deep learning based task systems normally rely on a large amount of manually labeled training data, which is expensive to obtain and subject to operator variations. Moreover, it does not always hold that the manually labeled data and the unlabeled data are sitting in the same distribution. In this paper, we alleviate these problems by proposing a discriminative consistent domain generation (\emph{DCDG}) approach to achieve a semi-supervised learning. The discriminative consistent domain is achieved by a double-sided domain adaptation. The double-sided domain adaptation aims to make a fusion of the feature spaces of labeled data and unlabeled data. In this way, we can fit the differences of various distributions between labeled data and unlabeled data. In order to keep the discriminativeness of generated consistent domain for the task learning, we apply an indirect learning for the double-sided domain adaptation. Based on the generated discriminative consistent domain, we can use the unlabeled data to learn the task model along with the labeled data via a consistent image generation. We demonstrate the performance of our proposed \emph{DCDG} on the late gadolinium enhancement cardiac MRI (LGE-CMRI) images acquired from patients with atrial fibrillation in two clinical centers for the segmentation of the left atrium anatomy (LA) and proximal pulmonary veins (PVs). The experiments show that our semi-supervised approach achieves compelling segmentation results, which can prove the robustness of \emph{DCDG} for the semi-supervised learning using the unlabeled data along with labeled data acquired from a single center or multicenter studies.
\end{abstract}

\section{Introduction}
Fitting the possible differences of distributions between labeled data and unlabeled data is of high importance for the semi-supervised learning. The usage of unlabeled data can overcome the limitation of insufficient labeled data, which is normally a hurdle in medical image analysis problems that lack labeled data. In practice, incorporating unlabeled data may fail due to the domain shift between labeled data and unlabeled data \cite{patel2015visual}.

The domain adaptation can learn generically adaptive representation domain but is subject to the limited discriminative feature domain for the task model learning. On the one hand, recent domain adaptation approaches usually introduce a discriminator to encourage data from one domain to generate a feature domain that is similar to the other one by keeping inter representation invariant between the two domains. They are based on a single adaptation direction. Thus the generated feature space is limited by one of the two domains. Because the original feature space of the other domain is lost, it can result in a reduced feature space of the two domains. On the other hand, the widespread domain adaptation approaches work with only the labeled data for the task model learning. The unlabeled data is only used to generate the domain adaptation space along with the labeled data based on an adversarial learning. Therefore, the discriminativeness of adapted features for subsequent task model still completely rely on the labeled data. Consequently, it is hard to guarantee the discriminativeness of adapted features for task learning of the unlabeled data. 
\begin{figure}[!hbtp]
\centering
\scalebox{.75}{
\includegraphics[width=1\textwidth]{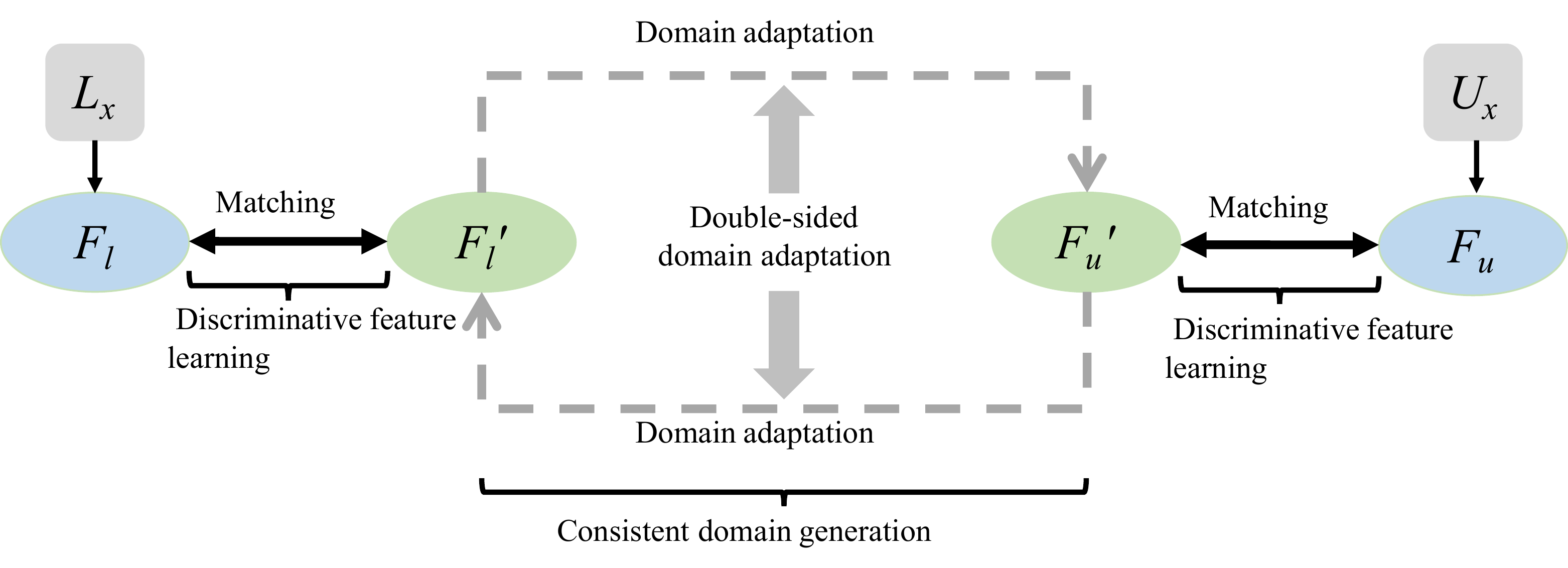}
}
\figcaption{\scriptsize{Indirect double-sided domain adaptation.The labeled data ($L_{x}$) and the unlabeled data ($U_{x}$) are mapped to the adapted consistent feature domain $F$($F_{l}$ and $F_{u}$). The $F$ is generated based on the double-sided domain adaptation via an indirect adversarial learning that the domain discriminator is applied to the feature domain $F'$, which is obtained from the predicted label. Feature matching is then used to keep the consistence between $F$ and $F'$.}} 
\label{fig:main}
\end{figure}

In this work, we propose a discriminative consistent domain generation method based on double-sided domain adaptation (as shown in Fig.\ref{fig:main}) to learn a task model (\emph{TM}) in a semi-supervised manner. In our proposed \emph{DCDG}, the available labeled data $L=(L_x,L_y)$ come from domain $D_{l}$ and the unlabeled data $U=(U_x)$ come from domain $D_u$. The $D_l$ and $D_u$ are from the same or a similar domain. We adopt the double-sided domain adaptation to generate a consistent feature domain $F$ that fuses the feature spaces of $D_l$ and $D_u$ instead of extracting the common parts of two domains or making one domain to adapt the other one. \emph{DCDG} shares a feature representation generator $FG\rightarrow F$ that maps $L_x$ and $U_x$ to the consistent feature domain $F$ $(F_{l}$ and $F_{u})$. For the purpose of discriminative feature generation, we map $F$ to the predicted label domain via $TM$. Then the domain discriminator maps predicted label domain to another feature domain ($F'$) to make an identification via indirect double-sided domain adaptation for $F$. During the indirect double-sided domain adaptation, the parameters of $TM$ are fixed and we constrain $F'$ to match the $F$ generated by $FG$. We can adapt the $F$ indirectly by adapting the $F'$ to guarantee the discriminativeness of feature domain $F$ for the subsequent learning of $TM$. During the discriminative consistent feature domain generation, we can learn the $TM$ although there are no available labels in $U_x$ for us to directly learn the $TM$. The $F'$ is matched with $F$, which is produced by both the $L_x$ and $U_x$. We can further map $F'$ to the generated labeled and unlabeled data ($L^{'}_x$ and $U^{'}_x$) to use the image consistency as the semi-supervised information to learn the $TM$.We demonstrate the performance of our proposed \emph{DCDG} for the left atrium segmentation \cite{chen2018multiview,yang2018multiview} on a  LGE-CMRI dataset, which plays an important
role in the management of atrial fibrillation and myocardial infarction\cite{yang2018fully,zhang2019deep}.

\section{Method}
The proposed $DCDG$ tries to generate the discriminative consistent feature domain with fused feature space from the labeled data $L_x$ and unlabeled data $U_x$ by the indirect double-sided domain adaptatipon. Meanwhile, we introduce an extra image consistent generation as the semi-supervised learning. Then the segmentation model can also be trained by the $U_x$ along with the $L_x$. Detailed network configuration can be found in the supplementary materials.
\begin{figure}[!hbtp]
\centering
\scalebox{0.85}{
\includegraphics[width=1\textwidth]{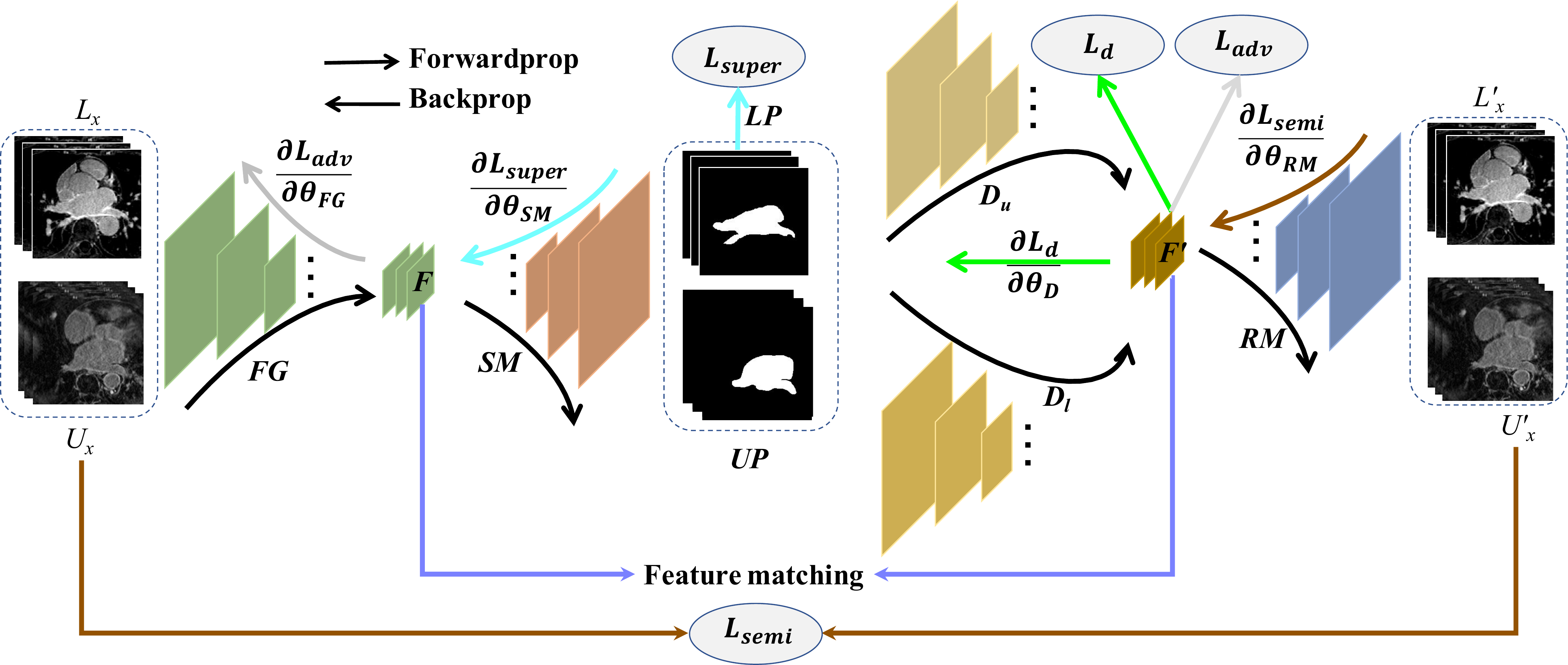}
}
\figcaption{\scriptsize{Illustration of the proposed $DCDG$. $L_{x}$ and $U_{x}$ represent the labeled data and unlabeled data respectively. They share a common feature generator ($FG$) to generate the consistent features $F$. Then the $F$ is mapped to the $LP$ and $UP$ of LA and PVs mask for $L_{x}$ and $U_{x}$ by the segmentation model ($SM$). Next, two domain discriminators with shared weights map the $LP$ and $UP$  to a feature domain ($F'$) to make an identification via indirect double-sided adversarial learning for $F$. 
During the adversarial learning, the parameters of $SM$ are fixed and we constrain $F'$ to match the $F$ generated by $FG$. Finally, the $F'$ is mapped to the $L'_{x}$ and $U'_{x}$ matched with $L_{x}$ and $U_{x}$.}}
\label{fig:network}
\end{figure}

\noindent\textbf{Discriminative feature extraction via indirect learning.}
We aim to generate the consistent feature domain via double-sided domain adaptation to learn a segmentation model using both the labeled data $L_x$ and the unlabeled data $U_x$. It is important to maintain the discriminativeness of feature representations for the generated consistent feature domain. Hence, we introduce the indirect learning for the discriminative feature domain extraction. In our proposed \emph{DCDG}, we use a feature generator $FG$ to generate the consistent feature domain $F$ ($F_{l}$ and $F_{u}$) from the $L_x$ and $U_x$ without using any knowledge of the source of images during testing \cite{zhang2018domain}. In order to guarantee the discriminativeness of the generated features for subsequent segmentation model, we introduce the indirect domain adaptation instead of direct domain adaptation for $F$. We map the generated feature $F$ to the estimated LA and PVs with probability maps ($LP$ and $UP$) of $L_x$ and $U_x$ via the segmentation model ($SM$). Then the domain discriminator ($D$) generates the feature domain $F'$ ($F'_{l}$ and $F'_{u}$) by the final convolutional layer from $LP$ and $UP$, and finally produces a scalar value to identify the $F'_{l}$ and $F'_{u}$. During the indirect learning, we fix the parameters of $SM$ and match the feature domains between $F'$ and $F$ via the squared $L_{2}$-norm that is defined as:
\begin{small}
\begin{equation}
\begin{split}
L_{fm}=||F_{l}-F'_{l}||^{2}_{2}+||F_{u}-F'_{u}||^{2}_{2}
\end{split}
\end{equation}
\end{small}
where the $F_{l}$ and $F_{u}$ is produce by the $FG$ $(L_{x})$ and $FG$ $(U_{x})$ respectively. the $F'_{l}$ and $F'_{u}$ is produce by the final convolutional layer of the domain discriminator.

Therefore, we identify the $F$ indirectly by identifying the $F'$ to guarantee the discriminativeness of feature domain $F$ for the learning of segmentation network. 

\vspace{0.3cm}
\noindent\textbf{Consistent feature domain generation via double-sided adaptation.}
In order to generate the consistent feature domain, we introduce two discriminators ($D_l$ and $D_u$) to achieve a double-sided domain adaptation which enables the features produced by the $L_x$ and $U_x$ to adapt each other as shown in Fig.\ref{fig:network}. $D_l$ is used to encourage the $L_x$ to generate the feature domain that is similar to the ones produced by $U_x$. $D_u$ is used to force the $U_x$ to generate the feature domain that is similar to the ones produced by $L_x$. Hence, a double-sided adversarial training is used to achieve the double-sided domain adaptation. During the double-sided adversarial training, for the learning of domain discriminator, $F'_{l}$ is used as the $fake$ feature and $F'_{u}$ is used as the $real$ feature to learn the $D_l$, while $F'_{l}$ is used as the $real$ feature and $F'_{u}$ is used as the $fake$ feature to learn the $D_u$ simultaneously. For the learning of feature generator $FG$, $D_l$ tries to identify the $F'_{l}$ as $real$ features and $D_u$ tries to identify the $F'_{u}$ as $real$ features. In order to reduce the parameters of network, we make the $D_l$ and $D_u$ to share a discriminator $D_{lu}$ to achieve the double adversarial learning directly. Inspired by this, we take the $F'_{l}$ and $F'_{u}$ as $real$ features and $false$ features respectively to directly learn the $D_{lu}$. When we learn the generator, the $F'_{l}$ and the $F'_{u}$ are assigned with the $False$ label and $True$ label respectively. Overall, during the double-sided adaptation, we fix the parameters of segmentation model $SM$ and aim to optimize the following $L_d$ and $L_{adv}$ for learning the domain discriminator and feature generator respectively: 
\begin{small}
\begin{equation}
L_{d}(\theta_{D_{lu}})=\sigma(D_{lu}(SM(FG(L_{x}))),1)+\sigma(D_{lu}(SM(FG(U_{x}))),0)
\end{equation}
\end{small}
\vspace{-0.2cm}
\begin{small}
\begin{equation}
L_{adv}(\theta_{FG})=\sigma(D_{lu}(SM(FG(L_{x}))),0)+\sigma(D_{lu}(SM(FG(U_{x}))),1)
\end{equation}
\end{small}
where the $L_{x}$ and $U_{x}$ represent the input of the labeled and unlabeled data, respectively. $\sigma$ is the binary cross-entropy loss. $SM$ represents the segmentation model. During the domain adaptation, the parameters of the $SM$ are fixed.

\vspace{0.3cm}
\noindent\textbf{Semi-supervised segmentation model learning.}
During the generation of the discriminative consistent feature domain, we learn the segmentation model $SM$ by  $L_{x}$ and $U_{x}$. Since there is no available labels for $U_{x}$, $U_{x}$ cannot be directly used for training the segmentation model along with the $L_{x}$. However, during the double-sided domain adaptation, we get the matched feature domains between $F$ and $F'$, while the $F$ is generated from the $L_{x}$ and $U_{x}$. Hence, we perform a reverse mapping ($RM$) that mapping the $F'$ to the $L'_{x}$ and $U'_{x}$ which are matched with $L_{x}$ and $U_{x}$ to achieve the consistent image generation. Then the $U_{x}$ can use a consistent  image loss as semi-supervised loss $L_{semi}$ to train the $SM$ along with the supervised loss $L_{super}$  from the $L_{x}$ . Finally, we use $L_{x}$ and $U_{x}$ to train $SM$ with the loss defined as follows:
\begin{equation}
L^{l}_{super,semi}(\theta_{SM},\theta_{RM},L_{x})=|| L_{x}-L'_{x}||^{2}_{2}+ DL(LP,L_{y}) 
\end{equation}
\begin{equation}
L^{u}_{semi}(\theta_{SM},\theta_{RM},U_{x})=||U_{x}-U'_{x}||^{2}_{2}
\end{equation}
where $LP$ is the estimated LA and PVs map. $L_{y}$ is the ground truth. $||\cdot||^{2}_{2}$ is the squared $L_{2}$-norm. $F$ and $F^{'}$ are the matched features. $DL(\cdot)$ represents the Dice loss function. We train the $SM$ after each epoch of the domain adaptation.

\vspace{0.3cm}
\noindent\textbf{Evaluation criteria.}
We use the region-based metrics: the Dice coefficient ($Dice=2\frac{|P\bigcap G|}{|P|+|G|}$) and intersection-over-union ($IoU=\frac{|P\bigcap G|}{|P|\bigcup|G|}$), which validate the predicted LA and PVs ($P$) against the ground-truth ($G$). And the surface-based metric of mean surface distance defined as $MSD=\frac{1}{2}[\overline{d}(S,S^\prime)+\overline{d}(S^\prime,S)]$, which $\overline{d}(S,S^\prime)$ is the mean of the distances between every surface voxel in  predicted mesh $S$ and the closest surface voxel in ground-truth mesh $S^\prime$.

\section{Experimental Results and Discussion}
In our experiments, we used two centers dataset (detailed imaging parameters can be found in the supplementary materials) for the LA and PVs segmentation to validate the proposed $DCDG$. The final segmentation model was obtained using `early stopping' on validation data. To demonstrate the performance of our proposed $DCDG$, we compare our proposed $DCDG$ with full supervised methods, e.g., 2\emph{D UNet} \cite{ronneberger2015u}, \emph{SegNet} \cite{badrinarayanan2017segnet} and a recent state-of-the-art 3D segmentation architecture namely 3\emph{D DenseNet} \cite{bui20173d}, and also make a comparision between $DCDG$ and a semi-supervised \cite{kozinski2017adversarial} ($AR$) method along with a domain adaptation \cite{tsai2018learning} ($ASOS$) method with the single-level adversarial learning. In addition, we also compared \emph{DCDG} to itself but used in a fully supervised manner, namely full supervised segmentation (\emph{FSS}).
\begin{table}[!hbtp]
 \setlength{\abovecaptionskip}{0pt} 
 \setlength{\belowcaptionskip}{0pt} 
 \caption{\scriptsize{Comparison of the performance of our proposed $DCDG$ on C1.}}
 \centering
 \scalebox{.5}{
 \setlength{\floatsep}{10pt plus 3pt minus 2pt} 
\begin{tabular}{C{3cm}C{4cm}C{4cm}C{4cm}}
\addlinespace
\toprule
Method & \multicolumn{1}{c}{IoU} & \multicolumn{1}{c}{MSD (mm)} &\multicolumn{1}{c}{Dice} \\ \midrule
  2D UNet & $0.8695\pm0.0294$ &$1.1149 \pm 0.4745$ & $0.9299\pm0.0170$ \cr 
 SegNet & $0.8418\pm0.0265$ &$1.2788\pm 0.2666$ & $0.9139\pm0.0157$ \cr
 3D DenseNet & $0.8414\pm0.0258$ & $1.4548\pm 0.5463$ & $0.9136\pm0.0153$ \cr
 FSS & $0.8719\pm0.0272$ & $1.0318\pm 0.2776$ & $0.9314\pm0.0157$\\ \midrule
 
 AR(25\%) & $0.8215\pm0.0330$ &$1.5829 \pm 0.4103$ & $0.9016\pm0.0200$ \cr 
 ASOS(25\%) & $0.8252\pm0.0312$ &$1.5645 \pm 0.4455$ & $0.9039\pm0.0189$ \cr 
 DCDG(25\%)  & $0.8474\pm0.0338$ &$1.2823 \pm 0.3524$ & $0.9170\pm0.0202$ \\ \midrule
 
 AR(50\%) & $0.8349\pm0.0377$ &$1.4538 \pm 0.4038$ & $0.9096\pm0.0288$ \cr 
 ASOS(50\%) & $0.8529\pm0.0341$ &$1.1741 \pm 0.3647$ & $0.9203\pm0.0184$ \cr 
 DCDG(50\%) & $0.8709\pm0.0276$ &$1.0236 \pm 0.2749$ & $0.9307\pm0.0161$ \\ \midrule
 
 AR(75\%) & $0.8541\pm0.0255$ &$1.1536 \pm 0.3531$ & $0.9211\pm0.0149$ \cr 
 ASOS(75\%) & $0.8602\pm0.0238$ &$1.1176 \pm 0.2805$ & $0.9247\pm0.0139$ \cr 
 DCDG(75\%) &$\mathbf{0.8803\pm0.0188}$ & $\mathbf{0.9285\pm0.2566}$ & $\mathbf{0.9362\pm0.0107}$\\ \bottomrule
\end{tabular}
 }
 \label{table:singleDomain}
\end{table}

\vspace{0.3cm}
\noindent\textbf{Experiments on a single center dataset.}
We performed multiple experiments on data with the same image domain based on different ratios of labeled data acquired at center 1 denoted as $C1$. The total number of $C1$ data is 175, in which 140 samples are randomly selected and used to train the model. We randomly selected 15 samples for model validation (7 pre-ablation and 8 post-ablation samples). We also randomly selected 20 samples for independent testing (10 pre-ablation and 10 post-ablation samples). During the experiment, we randomly selected different ratio r of labeled cases ($r=25\%, 50\%, 75\%$) from 140 samples along with the ratio (1-r) of the unlabeled data for the semi-supervised learning of \emph{DCDG} and \emph{AR}, while the \emph{ASOS} is learned by the labeled data of ratio r with the test data. The full supervised methods performed with the standard supervised training manner based on all the labeled data. The quantitative results are summarized in Table \ref{table:singleDomain}.  As shown in Table \ref{table:singleDomain}, when we use 50\% labeled data, the performance of \emph{DCDG} is superior to 2\emph{D UNet}, \emph{SegNet}, 3\emph{D DenseNet}. When we use 75\% labeled data, the performance of \emph{DCDG} is superior to \emph{FSS}, which use 100\% labeled data with fully supervised learning. Compared to those methods, we can use much less labeled data to obtain better results. It has great significance to avoid costly manual labeling when there is limited expert availability. Furthermore, compared with \emph{AR} and \emph{ASOS}, our proposed \emph{DCDG} also achieves the best results.
\begin{table}[!hbtp]
 \setlength{\abovecaptionskip}{0pt} 
 \setlength{\belowcaptionskip}{0pt} 
 \caption{\scriptsize{Comparison of the performance of our proposed $DCDG$ on C1 and C2.}}
 \centering
 \scalebox{.5}{
 \setlength{\floatsep}{10pt plus 3pt minus 2pt} 
\begin{tabular}{C{3cm}C{4cm}C{4cm}C{4cm}}
\addlinespace
\toprule
Method & \multicolumn{1}{c}{IoU} & \multicolumn{1}{c}{MSD (mm)} &\multicolumn{1}{c}{Dice} \\ \midrule
 2D UNet & $0.8125\pm0.0355$ &$1.7503 \pm 0.6119$ & $0.8961\pm0.0216$ \cr 
 SegNet & $0.7420\pm0.0474$ &$3.1581 \pm 1.4469$ & $0.8510\pm0.0310$ \cr 
 3D DenseNet & $0.7922\pm0.0339$ & $\mathbf{1.1819\pm 0.4967}$ & $0.8836\pm0.0212$  \cr
 AR & $0.6748\pm0.0610$ &$3.6165 \pm 1.0855$ & $0.8043\pm0.0434$ \cr 
 ASOS & $0.7878\pm0.0353$ & $1.8757\pm 0.5038$ & $0.8808\pm0.0222$ \\ \midrule
 DCDG &$\mathbf{0.8315\pm0.0254}$ & $1.4031\pm 0.4948$ & $\mathbf{0.9078\pm0.0151}$ \cr
 FSS & $0.8154\pm0.0401$ & $2.5606\pm 2.1998$ & $0.8978\pm0.0245$\\ \bottomrule
\end{tabular}
 }
 \label{table:twoDomain}
\end{table}

\begin{figure}[!hbtp]
\centering
\scalebox{0.8}{
\includegraphics[width=0.98\textwidth]{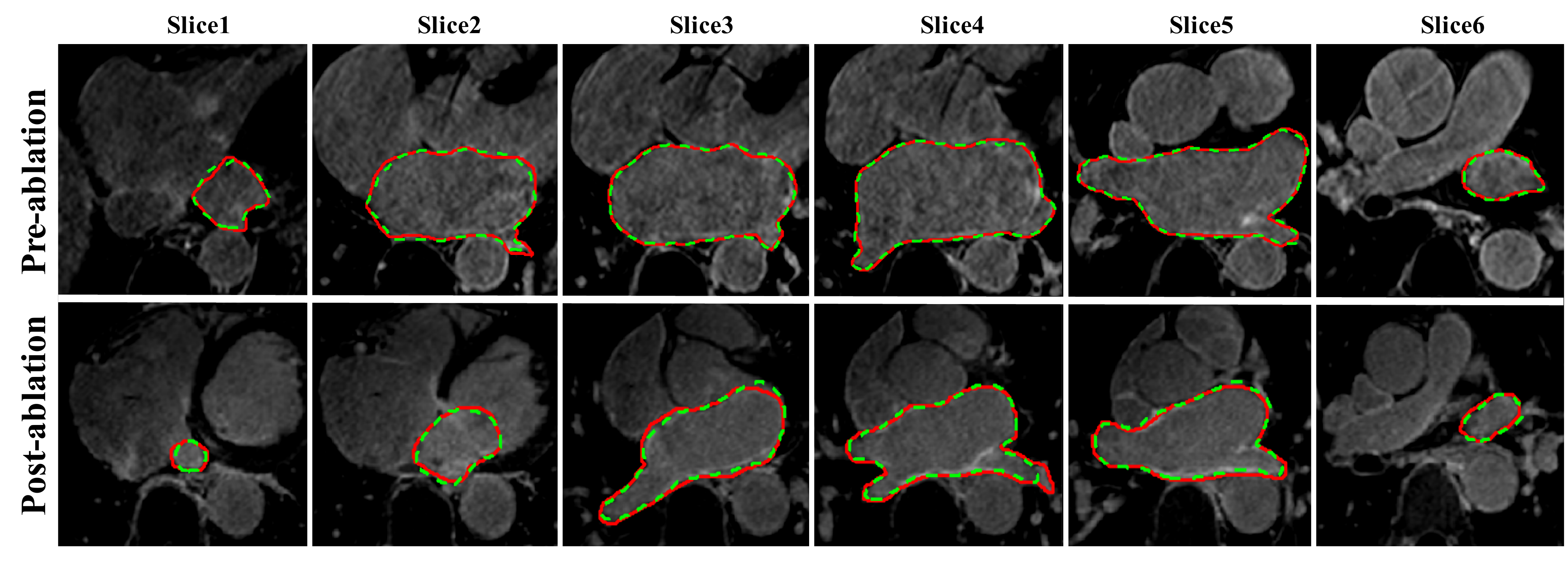}
}
\figcaption{\scriptsize{Qualitative visualization of LA and PVs segmentation results compared to the manual delineation for representative slices on a pre-ablation and a post-ablation 3D LGE-CMRI images. Each estimated segmentation is represented as a dashed green contour, and its corresponding manual delineation is represented as a red contour. }}
\label{fig:visualization}
\end{figure}

\noindent\textbf{Experiments on a two center dataset.}
For the experiment performed at $C1$ data was labelled as opposed to $C2$ which was unlabelled. The total number of $C2$ data is 94. We randomly selected 20 samples for testing (including 10 pre-ablation samples and 10 post-ablation samples). The remained 74 samples with no labels were used to train the \emph{DCDG} along with the 140 samples with available labels from $C1$. We also apply the same 15 validation dataset used in the learning from the single center data to validate the segmentation model during training.  For 2\emph{D UNet}, \emph{SegNet}, 3\emph{D DenseNet} and \emph{FSS}, supervised learning were performed on 140 samples with available labels from $C1$. As shown in Table \ref{table:twoDomain}, although  2\emph{D UNet}, \emph{SegNet}, 3\emph{D DenseNet} have achieved great success in many medical image segmentation applications, they can not be directly applied to learn a good model from the distribution of one domain to the other domain thus obtained sub-performance. In addition, our \emph{DCDG} also obtained the best results compared to the \emph{AR} and \emph{ASOS}. Fig.\ref{fig:visualization} visually gives a further illustration to demonstrate the good performance of \emph{DCDG}.
 \begin{figure}[!hbtp]
 \centering
 \setlength{\abovecaptionskip}{0pt} 
 \setlength{\belowcaptionskip}{0pt}
 \scalebox{.8}{
 \includegraphics[width=0.98\textwidth]{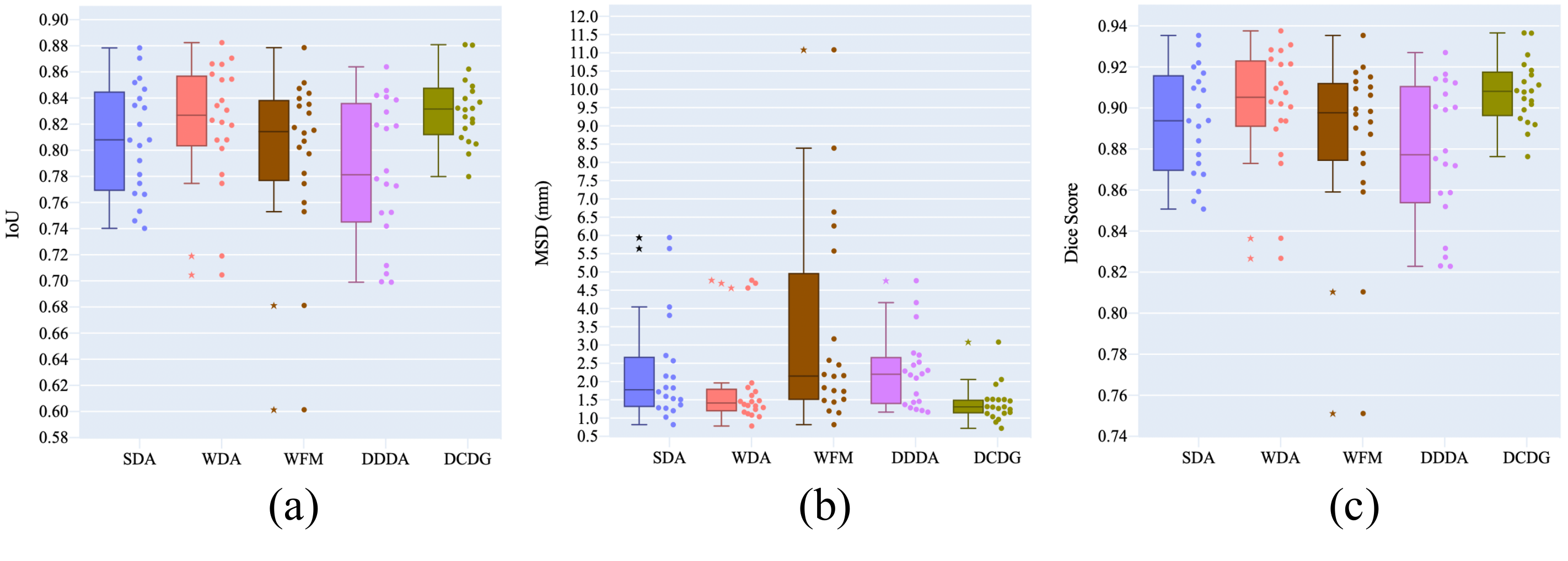}
 }
 \figcaption{\scriptsize{Boxplots of $IoU$,$MSD$ and $Dice$ evaluations for ablation tests.}}
 \label{fig:ablation}
 \end{figure}
 
\begin{figure}[!hbtp]
\centering
\scalebox{0.8}{
\includegraphics[width=0.99\textwidth]{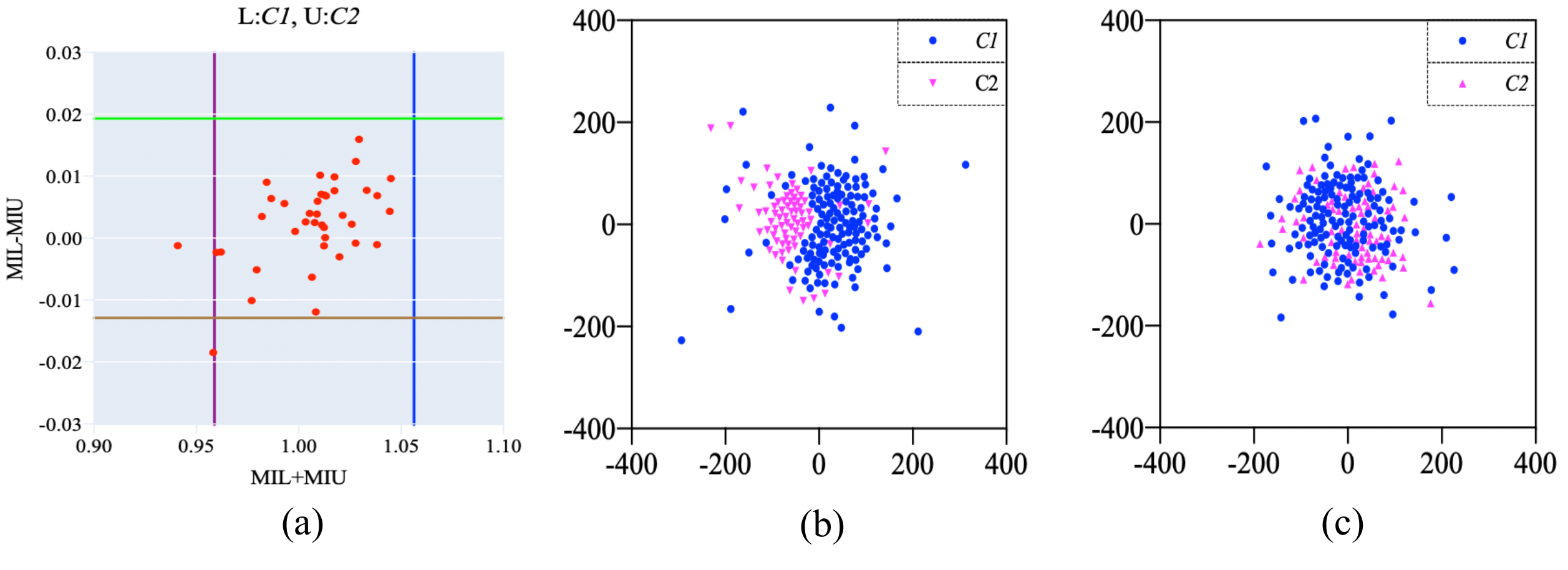}
}
\figcaption{\scriptsize{(a) The stability of double-sided domain adaptation. The $MIL$ and $MIU$ are the mean identification value of the discriminator on labeled samples and unlabeled samples respectively after each learning of the double-sided domain adaptation. The intervals between the two vertical lines and two horizontal lines are the confidence intervals of 95\% limits of agreement for MIL-MIU and MIL+MIU. (b) and (c): The t-Distributed Stochastic Neighbor Embedding (t-SNE) visualization of feature distribution on two-center data via double-sided adaptation (c) compared to the ones without adaptation(b). The red and blue points represent the samples from $C2$ and $C1$ respectively.}}
\label{fig:domainAdaptation}
\end{figure}

\noindent\textbf{Model variation study.}
To verify the effectiveness of each component in our proposed \emph{DCDG}, we perform Model variation study on $C1$ and $C2$. We take the structure of \emph{DCDG} as the baseline. Then we train the model with single domain adaptation (\emph{SDA}), without domain adaptation (\emph{WDA}), without feature matching (\emph{WFM}) and with the direct double-sided domain adaptation (\emph{DDDA}) for \emph{F}. We compare these models with the \emph{DCDG} and the results are shown in Fig.\ref{fig:ablation}. As shown in the Fig.\ref{fig:ablation}, our proposed \emph{DCDG} achieves the best results across these models on the measures of  \emph{IoU}, \emph{MSD} and \emph{Dice Score}, which proves the effectiveness of our \emph{DCDG}.

\vspace{0.3cm}
\noindent\textbf{Effectiveness of double-sided domain adaptation.}
In our proposed $DCDG$, the double-sided domain adaptation aims to generate the discriminative consistent feature domain. Its effectiveness can be demonstrated by the identified accuracy of discriminator for $L_{x}$ and $U_{x}$. Ideally, the probability values are identified by the discriminator for $L_{x}$ and $U_{x}$ are both 0.5. In our experiment, we record the $MIL$ and $MIU$ as shown in Fig.\ref{fig:domainAdaptation}(a). Ideally, the value of $MIL$ plus $MIU$ is 1, while the value of $MIL$ minus $MIU$ is 0. In this situation, the identification values of discriminator on labeled data and unlabeled data are both 0.5, which illustrates that the model achieves the double-sided domain adaptation. As shown in Fig.\ref{fig:domainAdaptation}(a), in each epoch during double-sided domain adaptation, the value of $MIL$ plus $MIU$ is close to 1. Meanwhile, the value of $MIL$ minus $MIU$ is close to 0. Furthermore, the adapted features show the fused features of $L_{x}$ and $U_{x}$ (Fig.\ref{fig:domainAdaptation}(c)) compared to no adapted feature distribution (Fig.\ref{fig:domainAdaptation}(b)) .

\section{Conclusion}
In this paper, we proposed a discriminative consistent domain generation for the semi-supervised learning. In our proposed $DCDG$, we investigate the double-sided domain adaptation based on an indirect adversarial learning to fit the differences between labeled data and unlabeled data and generate the discriminative feature domain with fused feature space. Validation of our framework has been performed against manually delineated ground truth of the LA and PVs segmentation task. Compared to other supervised, semi-supervisd and domain adaptation methods, our $DCDG$ has demonstrated superior performance. In conclusion, our proposed $DCDG$ makes it possible to create a robust semi-supervised learning model using the unlabeled data along with labeled data collected from a single center or a multicenter studies that can be well extended to solve other medical image analysis problems.

\section{Acknowledgments}
This work was supported in part by the Young Scientists Fund of the National Natural Science Foundation of China (61602003), in part by the Natural Science Foundation of China (61771464 and U1801265) and in part by the Guangdong Science and Technology (2018A050506031 and 2019B010110001).


\end{document}